\documentclass[runningheads]{llncs}
\usepackage{graphicx}
\usepackage{subcaption}
\usepackage{booktabs}

\usepackage{algorithm}
\usepackage{algpseudocode}

\usepackage{algpseudocode}
\usepackage{bbm}
\usepackage{amsmath}
\begin{document}

\title{Example-Based Explanations of\\Random Forest Predictions}
%
%
\author{Henrik Boström}
\authorrunning{H. Boström}
%
\institute{KTH Royal Institute of Technology, Stockholm, Sweden\\
\email{bostromh@kth.se}}
\maketitle              
\begin{abstract}

A random forest prediction can be computed by the scalar product of the labels of the training examples and a set of weights that are determined by the leafs of the forest into which the test object falls; each prediction can hence be explained exactly by the set of training examples for which the weights are non-zero. The number of examples used in such explanations is shown to vary with the dimensionality of the training set and hyperparameters of the random forest algorithm. This means that the number of examples involved in each prediction can to some extent be controlled by varying these parameters. However, for settings that lead to a required predictive performance, the number of examples involved in each prediction may be unreasonably large, preventing the user to grasp the explanations. In order to provide more useful explanations, a modified prediction procedure is proposed, which includes only the top-weighted examples. An investigation on regression and classification tasks shows that the number of examples used in each explanation can be substantially reduced while maintaining, or even improving, predictive performance compared to the standard prediction procedure.

\keywords{Random forests  \and Explainable machine learning \and Example-based explanations}

\end{abstract}

\section{Introduction}

Random forests \cite{breiman01} is a very popular and competitive machine learning algorithm that is widely considered to produce black-box models; even if each individual tree in a forest is interpretable, it is very hard to grasp an explanation that consist of several hundred (and sometimes even more) paths, each leading from the root of a tree to a leaf node, often also providing conflicting predictions. Techniques for explaining predictions of black-box models have received a lot of attention in recent years, with LIME \cite{ribeiro2016should} and SHAP \cite{lundberg2017unified} being two prominent examples of model-agnostic approaches that explain predictions by feature scores. In addition to explaining random forest predictions using feature scores, e.g., using TreeSHAP \cite{lundberg2020local}, techniques have also been proposed to approximate the random forests by interpretable rule sets, e.g., \cite{meinshausen2010node,bostrom2018explaining,ribeiro2018anchors,deng2019interpreting}. 

In contrast to explaining predictions by feature scores and rule sets, example-based explanation techniques explain the predictions by sets of examples \cite{molnar2022}. The latter can be useful in particular when the features are difficult to interpret. Such techniques do however require that the training examples can be presented to the user in an accessible way, e.g., as images. Apart from research on counterfactual explanation techniques \cite{guidotti2022counterfactual}, which synthesize new examples that lead to changing a prediction, example-based explanation techniques for tree-based methods have received limited attention. One exception is the prototype selection approach proposed in \cite{tan2020tree}, which applies clustering to find prototypical examples for each class to approximate a random forest by a nearest-neighbor procedure. In contrast to this approach and also to the previous rule-based approaches, we will in this work focus on exact (perfect fidelity) explanations; we hence do not rely on approximating the underlying model.

In \cite{meinshausen2006}, it was shown that a prediction of a random regression forest can be expressed as a scalar product of the labels of the training examples and a set of weights obtained from the leaf nodes into which the test object falls. In \cite{meinshausen2006}, the weights and labels were used to form cumulative distribution functions for quantile regression forests, while we will here instead consider them for explaining the predictions. As noted in \cite{geurts2006extremely}, the weight attribution also applies to classification; class membership of each training example can be encoded by a binary vector, which can be readily used when computing the scalar product. Using such a formulation, we can hence identify exactly which, and to what extent, training examples contribute to a prediction of a random forest for both classification and regression tasks. 

To the best of our knowledge, there has been no investigation of the effective number training examples used in the predictions of a random forest, i.e., the number of training examples with non-zero weights. This number may not only be dependent on the dimensionality of the training set, but also on the leaf and forest sizes. Even if we to some extent can control this number, potentially at the cost of reduced predictive performance, e.g., by keeping the number of trees in the forest small, we may still end up with a number of examples that is too large to be useful, e.g., interpreting hundreds of training examples may be as difficult as interpreting hundreds of paths. In this work, we propose to control this number by a modified prediction procedure; only the top-weighted training examples are used when forming the prediction for a test example. We hence end up with a procedure that is constrained in number (or weight) of the involved training examples, while providing an exact example-based explanation for each prediction, i.e., there is no approximation involved in how the actual prediction is computed. The main question that we will investigate is whether the effective number of examples can be reduced without sacrificing predictive performance.

In the next section, we describe the proposed approach in detail and in section~\ref{investigation}, we present results from an empirical investigation, where we first study how the way in which the random forest is formed may affect the effective number of training examples involved in the predictions, followed by an investigation, using both regression and classification datasets, of how controlling this number may impact the predictive performance. Finally, in Section~\ref{cr}, we discuss the main findings and outline some directions for future work.  

\section{Modifying the Prediction Procedure of Random Forests}
\label{approach}

We start out with some notation, before proceeding with the proposed modified prediction procedure of random forests.

\subsection{Random forests}

Each training example consists of an object and a label; let $X = \{\boldsymbol{x}_1, \ldots, \boldsymbol{x}_n\}$ denote the set of training objects and $y = \{y_1, \ldots, y_n\}$ the set of labels. For a regression problem, each $y_i \in \mathbbm{R}$. For a classification problem, where the class labels of the training objects are $\{y'_1, \ldots, y'_n\}$, with each $y'_i \in \{c_1, \ldots, c_k\}$, each label $y_i = \langle \mathbbm{1}(y'_i = c_1), \ldots, \mathbbm{1}(y'_i = c_k) \rangle$, i.e., a binary vector with zeros for all classes except the class label of the object.

Let $F = \{T_1, \ldots, T_s\}$ be a random forest; we refer to it as a classification forest if each $T_t$ is a classification tree, and a regression forest if each 
$T_t$ is a regression tree. Let $\hat{y_t} = T_t(\boldsymbol{x})$ denote the output (prediction) of a regression or classification tree $T_t$ for a (test) object $\boldsymbol{x}$; for a regression tree $\hat{y_t} \in \mathbbm{R}$ and for a classification tree $\hat{y_t} = \langle p_1, \ldots, p_k \rangle \in [0,1]^k$, such that $\sum_{i=1}^{k} p_i = 1$, i.e., the output is a class probability distribution. The prediction of the random forest $F$ for the test object $\boldsymbol{x}$ is:

\begin{equation}
   F(\boldsymbol{x}) = s^{-1}\sum_{t=1}^{s}T_t(\boldsymbol{x})
\end{equation}

Note that for a classification forest, the prediction is a class probability distribution, similar to the individual trees in the forest. Following \cite{meinshausen2006}, the above can be equivalently expressed as the scalar product of the labels of the training objects ($y$) and a set of weights $w_{\boldsymbol{x}} = \{w_{\boldsymbol{x}, 1}, \ldots, w_{\boldsymbol{x}, n}\}$:

\begin{equation}
   F(\boldsymbol{x}) = y \cdot w_{\boldsymbol{x}}
\end{equation}

\noindent where each weight $w_{\boldsymbol{x}, i}$ is defined by:

\begin{equation}
    w_{\boldsymbol{x}, i} = s^{-1}\sum_{t=1}^{s} w_{\boldsymbol{x}, i, t}
\end{equation}

\noindent and where $w_{\boldsymbol{x}, i, t}$ is defined by:

\begin{equation}
    w_{\boldsymbol{x}, i, t} = \frac{b_{\boldsymbol{x}, i, t}}{\sum_{j=1}^{n} b_{\boldsymbol{x}, j, t}}
\end{equation}


\noindent where $b_{\boldsymbol{x}, i, t}$ denotes the number of occurrences of training object $\boldsymbol{x}_i$ in the leaf node of the tree $T_t$ into which $\boldsymbol{x}$ falls. Note that in case a training object has not been part in the construction of a tree, i.e., it is out-of-bag for that tree, the corresponding weight will be zero independently of what leaf node the test object falls into. Note also that in case a training object does not occur in any of the leafs in any of the trees that the test object falls into, the total weight for the training example will be zero. 


\subsection{Modifying the predictions}

Without loss of generality, we may assume that the weights for a test object are sorted in decreasing order, i.e., we can form the random forest prediction by:

\begin{equation}
   F(\boldsymbol{x}) = \langle y_{\sigma_1}, \ldots, y_{\sigma_n} \rangle \cdot \langle w_{\boldsymbol{x}, \sigma_1}, \ldots, w_{\boldsymbol{x}, \sigma_n} \rangle
\end{equation}

\noindent where $w_{\boldsymbol{x}, \sigma_1}, \ldots, w_{\boldsymbol{x}, \sigma_n}$ are the weights for the test object ($\boldsymbol{x}$) sorted from the highest to the lowest with each $\sigma_i$ denoting the original index. We will investigate two alternative ways of making a prediction with a reduced number of training examples; 
by choosing the $k$ top-weighted objects only (Alg. \ref{alg:top-k}) and choosing a set of examples such that the cumulative weight exceeds a specified threshold (Alg. \ref{alg:cumulative}); for brevity, we denote each weight $w_{\boldsymbol{x}, i}$ by $w_i$ in the algorithms. Note that in both algorithms the selected weights need to be normalized.

\begin{minipage}{0.44\textwidth}
\begin{algorithm}[H]
    \centering
\caption{$k$ top-weighted}\label{alg:top-k}
\begin{algorithmic}[1]
\Require
\Statex $y = \{y_1, \ldots, y_n\}$
\Statex $w = \{w_1, \ldots, w_n\}$
\Statex $0 < k \leq n$ 
\State $\sigma_1, \ldots, \sigma_n \gets SortedIndex(w) $ 
\State $z \gets \sum_{i=1}^{k} w_{\sigma_i}$ 
\State $\hat{y} \gets $
\Statex $\langle y_{\sigma_1}, \ldots, y_{\sigma_k} \rangle \cdot \langle w_{\sigma_1}/z, \ldots, w_{\sigma_k}/z \rangle$ 
\\
\Return $\hat{y}$
\end{algorithmic}
\end{algorithm}
\end{minipage}
\hfill
\begin{minipage}{0.46\textwidth}
\begin{algorithm}[H]
    \centering
\caption{Cumulative weight}\label{alg:cumulative}
\begin{algorithmic}[1]
\Require 
\Statex $y = \{y_1, \ldots, y_n\}$
\Statex $w = \{w_1, \ldots, w_n\}$
\Statex $0 < c \leq 1$
\State $\sigma_1, \ldots, \sigma_n \gets SortedIndex(w) $ 
\State $z_j \gets \sum_{i=1}^{j} w_{\sigma_i}$, for $j = 1, \ldots, n$ 
\State $k \gets \min_{\{1, \ldots, n\}}$  \textrm{s.t.} $z_i \geq c$
\State $\hat{y} \gets $
\Statex $\langle y_{\sigma_1}, \ldots, y_{\sigma_k} \rangle \cdot \langle w_{\sigma_1}/z_k, \ldots, w_{\sigma_k}/z_k \rangle$
\\
\Return $\hat{y}$
\end{algorithmic}
\end{algorithm}
\end{minipage}

\section{Empirical Investigation}
\label{investigation}

In this section, we first investigate the effect of hyperparameter settings and dimensionality of the dataset on the effective number of training examples needed to form the predictions. We then present results from controlling the effective number of training examples on two prediction tasks.

\subsection{Observing the effective number of training examples}

\subsubsection{Experimental setup}

We have chosen the Lipophilicity dataset from MoleculeNet \cite{wu2018moleculenet}, which contains measurements of the octanol/water distribution coefficient for 4200 chemical compounds, represented by the simplified molecular-input line-entry system (SMILES). The Python package \texttt{RDKit}\footnote{https://www.rdkit.org} is used to generate features from the SMILES strings, more specifically, {\it Morgan fingerprints} (binary vectors, all of length 1024, if not stated otherwise). In addition to considering the original regression problem, we also frame it as a binary classification problem, where the task is to predict whether the target is greater than or equal to the mean of the targets, and as a multiclass classification problem, by equal-width binning of the regression values into ten categories.

We employ 10-fold cross validation, using the same folds and random seeds for all generated forests. In the first of four investigations for the three tasks, we vary the number of training examples by subsampling from the available training set, where a larger subsample always includes a smaller. In the second investigation, we vary the number of features by considering Morgan fingerprints of different sizes. In the third investigation, we vary the number of trees in the forests, and finally, in the fourth investigation, we vary the minimum sample size in each leaf. In addition to the average number of training examples that are assigned a non-zero weight for each test example (N), we also report the predictive performance; root mean squared error (RMSE) and Pearson correlation coefficient (Corr.) for regression, and accuracy (Acc.) and area under ROC curve (AUC) for classification.

The regression and classification forests are generated using \texttt{scikit-learn} \cite{scikit-learn}, with the default settings, except when stated otherwise. The methods to fit and apply the forests have been modified to allow for measuring and controlling the number of training examples used in the predictions. It has been verified that the generated predictions, when not limiting the number of involved training examples, are identical to those generated by the original implementation.

\subsubsection{Results for regression}

In Table~\ref{eff_no_lipo_reg}, the results from the four investigations on the regression task are shown. 
Table~\ref{eff_no_lipo_reg}a shows that the predictive performance is improved, as expected, when increasing the number of training examples. More interestingly, the effective number of training examples can be observed to decrease when increasing the training set size. A similar effect can be observed when increasing the number of features, as shown in Table~\ref{eff_no_lipo_reg}b; the predictive performance is improving while the effective number of training examples is decreasing. In contrast, increasing the number of trees in the forest leads to an increased number of training examples with non-zero weights, while the predictive performance is improved, albeit quite marginally, as seen in Table~\ref{eff_no_lipo_reg}c. Finally, Table~\ref{eff_no_lipo_reg}d shows that increasing the minimum leaf sample size has a detrimental effect on both predictive performance and the number of examples needed to explain the predictions, assuming that fewer examples are preferred. 

\begin{table}[h]

\begin{subtable}[h]{0.45\textwidth}
\centering
\begin{tabular}{r@{\hskip 0.25cm}|@{\hskip 0.25cm}r@{\hskip 0.25cm}r@{\hskip 0.25cm}r}
\toprule
\textbf{\#Ex.} & \textbf{RMSE} & \textbf{Corr.} & \textbf{N} \\
\midrule
500 & 1.042 & 0.503 & 61.7 \\
1000 & 0.977 & 0.587 & 59.8 \\
1500 & 0.940 & 0.630 & 58.8 \\
2000 & 0.904 & 0.665 & 57.6 \\
2500 & 0.894 & 0.674 & 56.2 \\
3000 & 0.876 & 0.689 & 55.5 \\
3500 & 0.860 & 0.703 & 54.1 \\
\bottomrule
\end{tabular}
\caption{No. of training examples}
\end{subtable}
\hspace{0.5cm}
\begin{subtable}[h]{0.45\textwidth}
\centering
\begin{tabular}{r@{\hskip 0.25cm}|@{\hskip 0.25cm}r@{\hskip 0.25cm}r@{\hskip 0.25cm}r}
\toprule
\textbf{\#Feat.} & \textbf{RMSE} & \textbf{Corr.} & \textbf{N} \\
\midrule
128 & 0.942 & 0.632 & 64.9 \\
256 & 0.901 & 0.671 & 62.1 \\
512 & 0.868 & 0.698 & 57.2 \\
1024 & 0.848 & 0.713 & 53.6 \\
2048 & 0.820 & 0.734 & 51.0 \\
4096 & 0.806 & 0.744 & 48.7 \\
8192 & 0.799 & 0.750 & 48.0 \\
\bottomrule
\end{tabular}
\caption{No. of features}
\end{subtable}
\newline
\vspace*{0.25cm}
\newline
\begin{subtable}[h]{0.45\textwidth}
\centering
\begin{tabular}{r@{\hskip 0.25cm}|@{\hskip 0.25cm}r@{\hskip 0.25cm}r@{\hskip 0.25cm}r}
\toprule
\textbf{\#Est.} & \textbf{RMSE} & \textbf{Corr.} & \textbf{N} \\
\midrule
100 & 0.852 & 0.710 & 53.6 \\
250 & 0.846 & 0.716 & 105.1 \\
500 & 0.846 & 0.716 & 167.4 \\
750 & 0.845 & 0.716 & 215.1 \\
1000 & 0.844 & 0.717 & 255.9 \\
1250 & 0.844 & 0.717 & 291.7 \\
1500 & 0.844 & 0.718 & 322.5 \\
\bottomrule
\end{tabular}
\caption{No. of trees}
\end{subtable}
\hspace{0.5cm}
\begin{subtable}[h]{0.45\textwidth}
\centering
\begin{tabular}{r@{\hskip 0.25cm}|@{\hskip 0.25cm}r@{\hskip 0.25cm}r@{\hskip 0.25cm}r}
\toprule
\textbf{\#Samp.} & \textbf{RMSE} & \textbf{Corr.} & \textbf{N} \\
\midrule
1 & 0.849 & 0.712 & 53.5 \\
5 & 0.872 & 0.698 & 267.2 \\
10 & 0.910 & 0.664 & 466.3 \\
15 & 0.938 & 0.638 & 632.0 \\
20 & 0.960 & 0.616 & 767.9 \\
25 & 0.977 & 0.598 & 879.3 \\
30 & 0.992 & 0.580 & 978.3 \\
\bottomrule
\end{tabular}
\caption{Minimum leaf sample size}
\end{subtable}
\newline
\newline
\caption{Regression results for the Lipophilicity dataset}
\label{eff_no_lipo_reg}
\end{table}

\subsubsection{Results for binary classification}

In Table~\ref{eff_no_lipo_binary}, the results for the binary classification task are shown. A first observation is that the number of training examples with non-zero weights are much larger for this task compared to the regression task; this can be attributed to the larger number of training examples falling into each leaf. In Table~\ref{eff_no_lipo_binary}a, the predictive performance is again observed to be improved with the number of training examples, but in contrast to the regression task, the effective number of training examples is consistently increasing with larger training sets. The picture is a bit different when increasing the number of features, as seen in Table~\ref{eff_no_lipo_binary}b; although the predictive performance is increasing with the number of features, the effective number of examples is instead changing non-monotonically, with a maximum reached at 1024 features. When it comes to increasing the number of trees in the forest, the results are similar to when considering the regression task; larger forests lead to larger number of training examples with non-zero weights, while the predictive performance (marginally) improves, as can be observed in Table~\ref{eff_no_lipo_binary}c. Finally, Table~\ref{eff_no_lipo_binary}d shows that increasing the minimum leaf sample size again has a detrimental effect on both predictive performance and the number of examples. 

\begin{table}[h]

\begin{subtable}[h]{0.45\textwidth}
\centering
\begin{tabular}{r@{\hskip 0.25cm}|@{\hskip 0.25cm}r@{\hskip 0.25cm}r@{\hskip 0.25cm}r}
\toprule
\textbf{\#Ex.} & \textbf{Acc.} & \textbf{AUC} & \textbf{N} \\
\midrule
500 & 0.683 & 0.747 & 306.8   \\
1000 & 0.718 & 0.789 & 402.4   \\
1500 & 0.744 & 0.819 & 455.2   \\
2000 & 0.751 & 0.828 & 493.1   \\
2500 & 0.768 & 0.843 & 521.0   \\
3000 & 0.774 & 0.853 & 545.7   \\
3500 & 0.772 & 0.857 & 560.5   \\
\bottomrule
\end{tabular}
\caption{No. of training examples}
\end{subtable}
\hfill
\begin{subtable}[h]{0.45\textwidth}
\centering
\begin{tabular}{r@{\hskip 0.25cm}|@{\hskip 0.25cm}r@{\hskip 0.25cm}r@{\hskip 0.25cm}r}
\toprule
\textbf{\#Feat.} & \textbf{Acc.} & \textbf{AUC} & \textbf{N} \\
\midrule
128 & 0.760 & 0.837 & 388.5 \\
256 & 0.770 & 0.849 & 473.3 \\
512 & 0.772 & 0.856 & 530.1 \\
1024 & 0.785 & 0.863 & 565.5 \\
2048 & 0.782 & 0.868 & 540.5 \\
4096 & 0.791 & 0.874 & 489.7 \\
8192 & 0.797 & 0.876 & 392.2 \\
\bottomrule
\end{tabular}
\caption{No. of features}
\end{subtable}
\newline
\vspace*{0.25cm}
\newline
\begin{subtable}[h]{0.45\textwidth}
\centering
\begin{tabular}{r@{\hskip 0.25cm}|@{\hskip 0.25cm}r@{\hskip 0.25cm}r@{\hskip 0.25cm}r}
\toprule
\textbf{\#Est.} & \textbf{Acc.} & \textbf{AUC} & \textbf{N} \\
\midrule
100 & 0.789 & 0.865 & 566.3   \\
250 & 0.786 & 0.865 & 983.8   \\
500 & 0.788 & 0.867 & 1382.4   \\
750 & 0.791 & 0.867 & 1624.7   \\
1000 & 0.792 & 0.868 & 1807.5   \\
1250 & 0.790 & 0.869 & 1939.0   \\
1500 & 0.791 & 0.868 & 2050.3   \\
\bottomrule
\end{tabular}
\caption{No. of trees}
\end{subtable}
\hspace{0.5cm}
\begin{subtable}[h]{0.45\textwidth}
\centering
\begin{tabular}{r@{\hskip 0.25cm}|@{\hskip 0.25cm}r@{\hskip 0.25cm}r@{\hskip 0.25cm}r}
\toprule
\textbf{\#Samp.} & \textbf{Acc.} & \textbf{AUC} & \textbf{N} \\
\midrule
1 & 0.789 & 0.864 & 567.6 \\
5 & 0.766 & 0.849 & 702.3 \\
10 & 0.752 & 0.830 & 1295.9 \\
15 & 0.740 & 0.816 & 1759.9 \\
20 & 0.730 & 0.807 & 2135.4 \\
25 & 0.721 & 0.800 & 2423.2 \\
30 & 0.710 & 0.791 & 2668.6 \\
\bottomrule
\end{tabular}
\caption{Minimum leaf sample size}
\end{subtable}
\newline
\newline
\caption{Binary classification results for the Lipophilicity dataset}
\label{eff_no_lipo_binary}
\end{table}

\subsubsection{Results for multiclass classification}

In Table~\ref{eff_no_lipo_multi}, the results for the multiclass classification task are shown. The effective number of training examples used in the predictions can be observed to fall in between of regression forests and binary classification forests. Due to the more fine-grained class labels, the tree growth typically continues beyond that of the binary classification trees, resulting in leafs with fewer examples, which has a direct effect on the number of training examples with non-zero weights. Table~\ref{eff_no_lipo_multi}a shows that the predictive performance improves when increasing the number of training examples, as observed also for the previous tasks, but in contrast to these, the effective number of training examples is not monotonically increasing or decreasing with larger training sets, but peaks near the middle of the considered range of training set sizes. Again, the predictive performance is increasing with the number of features, and similarly to the regression task, but different from the binary classification task, the effective number of involved training examples is decreasing with the dimensionality, as seen in Table~\ref{eff_no_lipo_multi}b. As was observed for both the regression and binary classification tasks, larger forests consistently lead to increasing the effective number of used examples, while the predictive performance is marginally affected, as can be observed in Table~\ref{eff_no_lipo_multi}c. Finally, Table~\ref{eff_no_lipo_multi}d shows that similar to the previous two cases, an increased minimum leaf sample size results in lower predictive performance and larger number of examples. 

\begin{table}[h]

\begin{subtable}[h]{0.45\textwidth}
\centering
\begin{tabular}{r@{\hskip 0.25cm}|@{\hskip 0.25cm}r@{\hskip 0.25cm}r@{\hskip 0.25cm}r}
\toprule
\textbf{\#Ex.} & \textbf{Acc.} & \textbf{AUC} & \textbf{N} \\
\midrule
500 & 0.238 & 0.604 & 125.9   \\
1000 & 0.257 & 0.637 & 136.0   \\
1500 & 0.283 & 0.664 & 134.7   \\
2000 & 0.292 & 0.679 & 134.7   \\
2500 & 0.297 & 0.687 & 130.1   \\
3000 & 0.303 & 0.701 & 128.1   \\
3500 & 0.309 & 0.705 & 126.0   \\
\bottomrule
\end{tabular}
\caption{No. of training examples}
\end{subtable}
\hspace{0.5cm}
\begin{subtable}[h]{0.45\textwidth}
\centering
\begin{tabular}{r@{\hskip 0.25cm}|@{\hskip 0.25cm}r@{\hskip 0.25cm}r@{\hskip 0.25cm}r}
\toprule
\textbf{\#Feat.} & \textbf{Acc.} & \textbf{AUC} & \textbf{N} \\
\midrule
128 & 0.311 & 0.694 & 132.1 \\
256 & 0.310 & 0.704 & 138.5 \\
512 & 0.320 & 0.708 & 134.3 \\
1024 & 0.310 & 0.714 & 124.8 \\
2048 & 0.320 & 0.719 & 107.8 \\
4096 & 0.315 & 0.723 & 92.2 \\
8192 & 0.324 & 0.722 & 77.7 \\
\bottomrule
\end{tabular}
\caption{No. of features}
\end{subtable}
\newline
\vspace*{0.25cm}
\newline
\begin{subtable}[h]{0.45\textwidth}
\centering
\begin{tabular}{r@{\hskip 0.25cm}|@{\hskip 0.25cm}r@{\hskip 0.25cm}r@{\hskip 0.25cm}r}
\toprule
\textbf{\#Est.} & \textbf{Acc.} & \textbf{AUC} & \textbf{N} \\
\midrule
100 & 0.318 & 0.715 & 123.7   \\
250 & 0.320 & 0.723 & 256.2   \\
500 & 0.316 & 0.724 & 421.5   \\
750 & 0.313 & 0.725 & 549.9   \\
1000 & 0.318 & 0.725 & 654.2   \\
1250 & 0.319 & 0.727 & 743.3   \\
1500 & 0.316 & 0.726 & 820.7   \\
\bottomrule
\end{tabular}
\caption{No. of trees}
\end{subtable}
\hspace{0.5cm}
\begin{subtable}[h]{0.45\textwidth}
\centering
\begin{tabular}{r@{\hskip 0.25cm}|@{\hskip 0.25cm}r@{\hskip 0.25cm}r@{\hskip 0.25cm}r}
\toprule
\textbf{\#Samp.} & \textbf{Acc.} & \textbf{AUC} & \textbf{N} \\
\midrule
1 & 0.321 & 0.715 & 124.0 \\
5 & 0.291 & 0.722 & 698.0 \\
10 & 0.278 & 0.713 & 1308.7 \\
15 & 0.268 & 0.704 & 1782.8 \\
20 & 0.261 & 0.698 & 2176.4 \\
25 & 0.261 & 0.692 & 2436.8 \\
30 & 0.250 & 0.687 & 2694.8 \\
\bottomrule
\end{tabular}
\caption{Minimum leaf sample size}
\end{subtable}
\newline
\newline
\caption{Multiclass classification results for the Lipophilicity dataset}
\label{eff_no_lipo_multi}
\end{table}

\subsubsection{Summary of the findings}

Two consistent patterns were observed across the three considered predictions tasks; increasing the number of trees in the forests leads to improved predictive performance and an increased number of training examples involved in the predictions, while increasing the minimum leaf sample size leads to deteriorated predictive performance and a substantial increase in the number of training examples with non-zero weight. The last finding suggests that the smallest possible minimum leaf sample size, i.e., 1, should be employed, which indeed is the default for random forests in \texttt{scikit-learn}. When it comes to the number of trees in the forests, there is a trade-off between the predictive performance and the effective number of examples; there may be reasons to use more than the default of 100 trees in \texttt{scikit-learn}, but the relatively small improvements beyond 500 trees or so come at a quite substantial cost in the number of examples needed to explain the predictions.

The most surprising finding was that increasing the training set size may not only lead to improved predictive performance, as expected, but also to a reduced number of training examples used in the predictions, as was observed for the regression and multiclass classification tasks. This means that reducing the training set is not always a good strategy to minimize the effective number of examples. A similar finding was made with respect to the number of features; a higher dimensionality consistently lead to higher performance, and for the regression and multiclass classification tasks, the lowest number of examples were used when the highest number of features were considered. The two last findings suggest that using as many training examples and features as possible can, at least in some cases, be advisable as both the predictive performance and the number of training examples used in the explanations benefit from this.

\subsection{Controlling the number of examples used in the predictions}

\subsubsection{Results for regression}
As in the previous section, we here consider the Lipophilicity dataset for the (original) regression task, using the largest number of features (8192). Again, we perform 10-fold cross-validation, here using a regression forest with 500 trees and with all other parameters set to the default.  

In Table~\ref{control_lipo_reg}, the results from controlling the effective number of examples (Table~\ref{control_lipo_reg}a) and the cumulative weight of the examples  (Table~\ref{control_lipo_reg}b) are presented. In the column \textbf{N}, the effective number of training examples are shown; note that in the first sub-table, this number may be less than the specified number (\textbf{k}), in particular for large values of the latter, as the number of examples receiving a non-zero weight may be less than \textbf{k}. The column \textbf{W} presents the average observed cumulative weight of the examples;  note that in the second sub-table, this number is typically larger than the specified cumulative weight (\textbf{c}), in particular for smaller values of \textbf{c}, as the latter provides a lower bound. The predictive performance of the standard regression forest is shown in the last row of Table~\ref{control_lipo_reg}b (where \textbf{c} = 1.0), where on average 137.5 training examples receive a non-zero weight. The results in Table~\ref{control_lipo_reg}a shows that the same predictive performance as the original forest can be obtained with as few as 15-20 training examples, which corresponds to a reduction of 85-90\% in the number of examples needed to explain the predictions. Interestingly, it can be observed in Table~\ref{control_lipo_reg}b that using a cumulative weight of 0.7 outperforms the original regression forest (as well as most other considered settings for the cumulative weight), while reducing the number of involved examples to less than a fourth.

\begin{figure}

\begin{subfigure}{0.3\textwidth}
    \centering
    \includegraphics[trim={0 2.75cm 0 0.25cm},clip, scale=0.42]{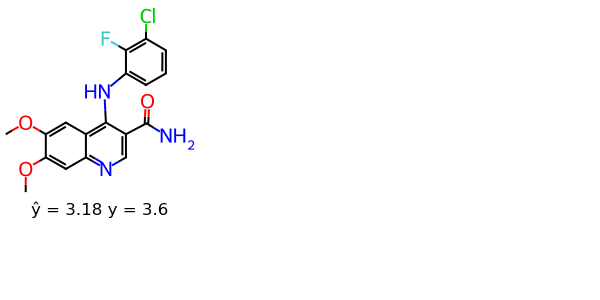}
    \caption{Test example}
\end{subfigure}
\hfill
\begin{subfigure}{0.3\textwidth}
    \centering
    \includegraphics[trim={0 0 0 0.25cm},clip, scale=0.45]{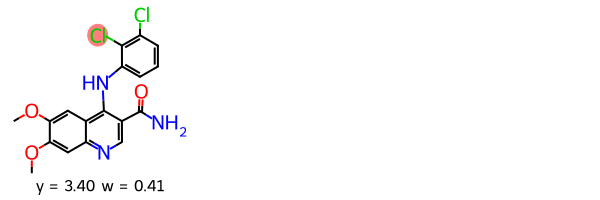}
    \caption{1st example}
\end{subfigure}
\hfill
\begin{subfigure}{0.3\textwidth}
    \centering
    \includegraphics[trim={0 0 0 0.25cm},clip, scale=0.45]{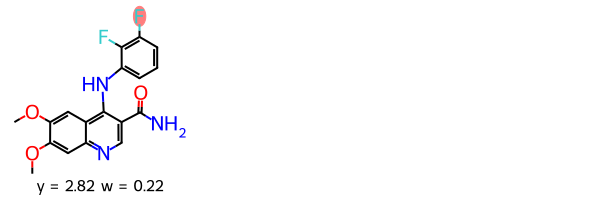}
    \caption{2nd example}
\end{subfigure}
\newline
\newline
\begin{subfigure}{0.3\textwidth}
    \centering
    \includegraphics[trim={0 0 0 0.25cm},clip, scale=0.45]{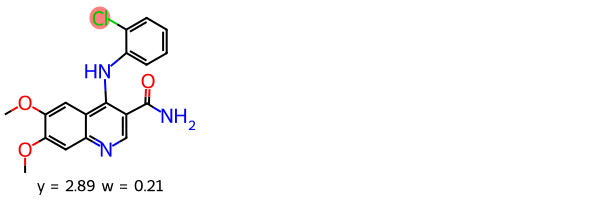}
    \caption{3rd example}
\end{subfigure}
\hfill
\begin{subfigure}{0.3\textwidth}
    \centering
    \includegraphics[trim={0 0 0 0.25cm},clip, scale=0.45]{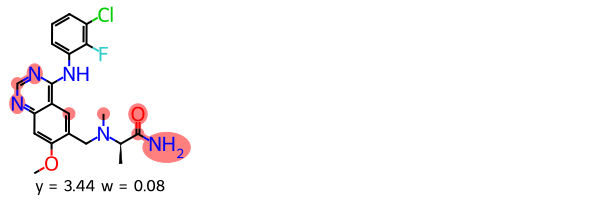}
    \caption{4th example}
\end{subfigure}
\hfill
\begin{subfigure}{0.3\textwidth}
    \centering
    \includegraphics[trim={0 0 0 0.25cm},clip, scale=0.45]{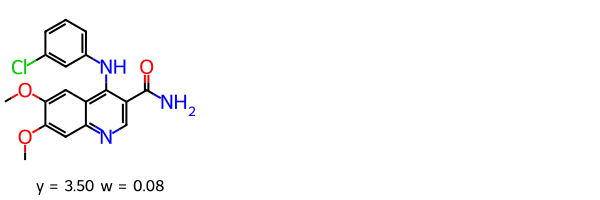}
    \caption{5th example}
\end{subfigure}

\caption{Test example and $k = 5$ training examples}
\label{fig:molecules}
\end{figure}

In Fig.~\ref{fig:molecules}, we illustrate the use of the above model trained on 90\% of the data and applied to a random test object, using the top five ($k = 5$) training examples for forming the predictions. Below the test object in Fig.~\ref{fig:molecules}a, the predicted ($\hat{y}$) and actual ($y$) values are shown. Below each of the training objects in Fig.~\ref{fig:molecules}b-f, the label ($y$) and the weight ($w$) are shown. Highlighted atoms in the training objects indicate parts that are missing in the test object. Even without knowledge about the particular features used by the black-box model, the user can inspect and reason about the actual objects that constitute the basis for the prediction. 

\begin{table}[h]

\begin{subtable}[h]{0.45\textwidth}
\centering

\begin{tabular}{r@{\hskip 0.25cm}|@{\hskip 0.25cm}r@{\hskip 0.25cm}r@{\hskip 0.25cm}r@{\hskip 0.25cm}r}
\toprule
\textbf{k} & \textbf{N} & \textbf{W} & \textbf{RMSE} & \textbf{Corr.} \\
\midrule
1 & 1.0 & 0.222 & 0.988 & 0.656 \\
3 & 3.0 & 0.398 & 0.850 & 0.723 \\
5 & 5.0 & 0.481 & 0.822 & 0.737 \\
10 & 10.0 & 0.586 & 0.803 & 0.747 \\
15 & 15.0 & 0.644 & 0.796 & 0.751 \\
20 & 20.0 & 0.685 & 0.793 & 0.752 \\
30 & 29.9 & 0.743 & 0.789 & 0.755 \\
50 & 48.6 & 0.818 & 0.788 & 0.756 \\
100 & 87.5 & 0.915 & 0.791 & 0.755 \\
\bottomrule
\end{tabular}
\caption{Varying number of examples}
\end{subtable}
\hspace{0.5cm}
\begin{subtable}[h]{0.45\textwidth}
\centering

\begin{tabular}{r@{\hskip 0.25cm}|@{\hskip 0.25cm}r@{\hskip 0.25cm}r@{\hskip 0.25cm}r@{\hskip 0.25cm}r}
\toprule
\textbf{c} & \textbf{W} & \textbf{N} & \textbf{RMSE} & \textbf{Corr.} \\
\midrule
0.1 & 0.238 & 1.4 & 0.918 & 0.693 \\
0.2 & 0.289 & 2.6 & 0.865 & 0.717 \\
0.3 & 0.365 & 4.7 & 0.834 & 0.732 \\
0.4 & 0.447 & 7.8 & 0.817 & 0.740 \\
0.5 & 0.535 & 12.6 & 0.804 & 0.747 \\
0.6 & 0.621 & 19.9 & 0.796 & 0.751 \\
0.7 & 0.713 & 31.3 & 0.792 & 0.753 \\
0.8 & 0.806 & 50.7 & 0.792 & 0.753 \\
0.9 & 0.902 & 82.9 & 0.793 & 0.753 \\
1.0 & 1.000 & 137.5 & 0.796 & 0.752 \\
\bottomrule
\end{tabular}

\caption{Varying cumulative weight}
\end{subtable}
\newline
\newline
\caption{Regression results for the Lipophilicity dataset}
\label{control_lipo_reg}
\end{table}

\subsubsection{Results for classification}
We here consider the MNIST dataset with 70~000 handwritten digits, represented by 784 features (28x28 pixel boxes) and for which the set of class labels is $\{0, \ldots, 9\}$. We employ ten-fold cross-validation and consider classification forests of 500 trees with all other parameters set to default.

In Table~\ref{control_mnist}, the results from controlling the effective number of examples (\textbf{k} in Table~\ref{control_mnist}a) and the cumulative weight of the examples  (\textbf{c} in Table~\ref{control_mnist}b) are presented, again with the columns \textbf{N} and \textbf{W} corresponding to the effective number and the cumulative weight of the examples, respectively. The predictive performance of the standard random forest is shown in the last row of Table~\ref{control_mnist}b (where \textbf{c} = 1.0), where on average 5846.9 training examples receive a non-zero weight. The results in Table~\ref{control_mnist}a shows that the original forest can be outperformed with as few as $k = 10$ training examples; this corresponds to a reduction of 99.8\% in the number of examples needed to explain the predictions. Similar results can be observed for several of the settings in Table~\ref{control_mnist}b.

\begin{figure}

\begin{subfigure}{0.3\textwidth}
    \centering
    \includegraphics[trim={0 0 0 1.9cm},clip, scale=0.25]{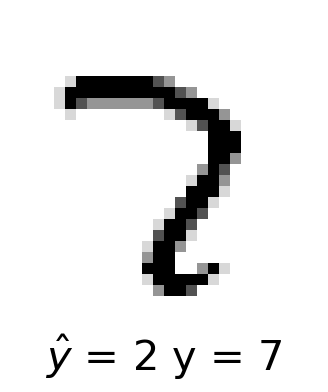}
    \caption{Test example}
\end{subfigure}
\hfill
\begin{subfigure}{0.3\textwidth}
    \centering
    \includegraphics[trim={0 0 0 1.9cm},clip, scale=0.25]{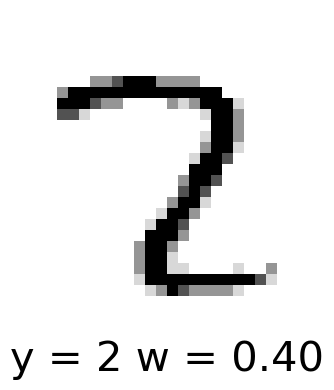}
    \caption{1st example}
\end{subfigure}
\hfill
\begin{subfigure}{0.3\textwidth}
    \centering
    \includegraphics[trim={0 0 0 1.9cm},clip, scale=0.25]{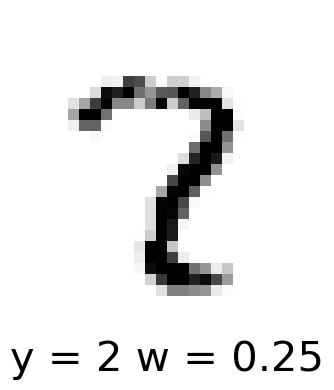}
    \caption{2nd example}
\end{subfigure}
\newline
\newline
\begin{subfigure}{0.3\textwidth}
    \centering
    \includegraphics[trim={0 0 0 1.9cm},clip, scale=0.25]{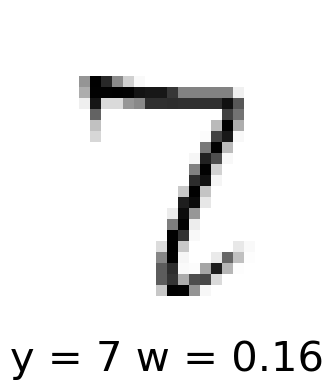}
    \caption{3rd example}
\end{subfigure}
\hfill
\begin{subfigure}{0.3\textwidth}
    \centering
    \includegraphics[trim={0 0 0 1.9cm},clip, scale=0.25]{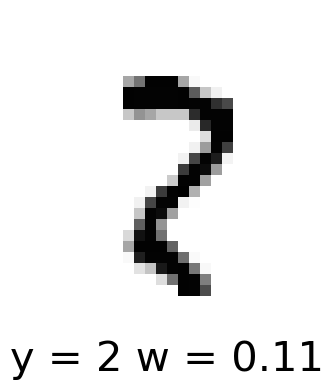}
    \caption{4th example}
\end{subfigure}
\hfill
\begin{subfigure}{0.3\textwidth}
    \centering
    \includegraphics[trim={0 0 0 1.9cm},clip, scale=0.25]{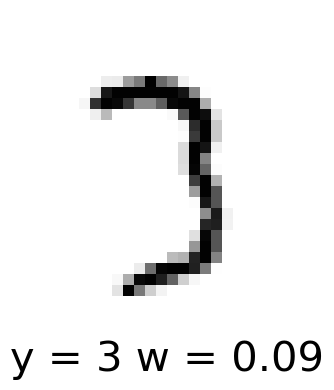}
    \caption{5th example}
\end{subfigure}

\caption{Test example and $k = 5$ training examples}
\label{fig:mnist}
\end{figure}

In Fig.~\ref{fig:mnist}, we illustrate the use of a classification forest trained on 90\% of the data when forming predictions using the top five ($k = 5$) training examples. We have randomly selected one test object with label $y = 7$ incorrectly predicted as $\hat{y} = 2$, shown in Fig.~\ref{fig:mnist}a; the predicted label is chosen according to the predicted class probability distribution $\langle 0, 0, 0.75, 0.09, 0, 0, 0, 0.16, 0, 0 \rangle$ (over the labels $0, \ldots, 9$), which is defined by the weights ($w$) and labels ($y$) of the training objects in Fig.~\ref{fig:mnist}b-f. Again, the user may inspect the training examples that fully explain the prediction, i.e., no other examples are involved in forming it, and e.g., reason about whether the prediction is reliable or not.

\begin{table}[h]

\begin{subtable}[h]{0.45\textwidth}
\centering
\begin{tabular}{r@{\hskip 0.25cm}|@{\hskip 0.25cm}r@{\hskip 0.25cm}r@{\hskip 0.25cm}r@{\hskip 0.25cm}r}
\toprule
\textbf{k} & \textbf{N} & \textbf{W} & \textbf{Acc.} & \textbf{AUC} \\
\midrule
1 & 1.0 & 0.008 & 0.861 & 0.923 \\
3 & 3.0 & 0.020 & 0.922 & 0.988 \\
5 & 5.0 & 0.029 & 0.952 & 0.996 \\
10 & 10.0 & 0.048 & 0.974 & 0.999 \\
15 & 15.0 & 0.064 & 0.980 & 0.999 \\
20 & 20.0 & 0.077 & 0.982 & 0.999 \\
30 & 30.0 & 0.101 & 0.983 & 1.000 \\
50 & 50.0 & 0.140 & 0.984 & 1.000 \\
100 & 100.0 & 0.217 & 0.984 & 1.000 \\
\bottomrule
\end{tabular}
\caption{Varying number of examples}
\end{subtable}
\hspace{0.5cm}
\begin{subtable}[h]{0.45\textwidth}
\centering
\begin{tabular}{r@{\hskip 0.25cm}|@{\hskip 0.25cm}r@{\hskip 0.25cm}r@{\hskip 0.25cm}r@{\hskip 0.25cm}r}
\toprule
\textbf{c} & \textbf{W} & \textbf{N} & \textbf{Acc.} & \textbf{AUC} \\
\midrule
0.1 & 0.101 & 51.7 & 0.983 & 0.999 \\
0.2 & 0.201 & 136.8 & 0.984 & 1.000 \\
0.3 & 0.301 & 252.4 & 0.984 & 1.000 \\
0.4 & 0.400 & 403.4 & 0.983 & 1.000 \\
0.5 & 0.500 & 600.3 & 0.982 & 1.000 \\
0.6 & 0.600 & 859.4 & 0.980 & 0.999 \\
0.7 & 0.700 & 1209.9 & 0.979 & 0.999 \\
0.8 & 0.800 & 1712.8 & 0.976 & 0.999 \\
0.9 & 0.900 & 2537.9 & 0.975 & 0.999 \\
1.0 & 1.000 & 5846.9 & 0.972 & 0.999 \\
\bottomrule
\end{tabular}
\caption{Varying cumulative weight}
\end{subtable}
\newline
\newline
\caption{Classification results for the MNIST dataset}
\label{control_mnist}
\end{table}

\section{Concluding remarks}
\label{cr}

An investigation of the number of training examples involved in random forest predictions has been presented, highlighting the impact of dataset properties and hyperparameter settings. An approach to controlling this number by including only the top-weighted examples has been proposed, and an empirical investigation shows that this approach may substantially reduce the effective number of training examples involved in the predictions, while maintaining, and even improving, the predictive performance compared to the standard prediction procedure. 

Directions for future research include extending the empirical investigation, e.g., by considering more datasets and hyperparameter settings, and investigating other approaches to selecting examples based on the weights. 
Other directions concern studying the usability of the example-based explanations when solving practical tasks, and also exploring combinations of explanation techniques, e.g., complementing the example-based explanations with rules or feature scores.

\bibliography{references}

\begin{thebibliography}{10}
\providecommand{\url}[1]{\texttt{#1}}
\providecommand{\urlprefix}{URL }
\providecommand{\doi}[1]{https://doi.org/#1}

\bibitem{bostrom2018explaining}
Bostr{\"o}m, H., Gurung, R.B., Lindgren, T., Johansson, U.: Explaining random forest predictions with association rules. Archives of Data Science, Series A (Online First)  \textbf{5}(1), ~A05 (2018)

\bibitem{breiman01}
Breiman, L.: Random forests. Machine Learning  \textbf{45}(1),  5--32 (2001)

\bibitem{deng2019interpreting}
Deng, H.: Interpreting tree ensembles with intrees. International Journal of Data Science and Analytics  \textbf{7}(4),  277--287 (2019)

\bibitem{geurts2006extremely}
Geurts, P., Ernst, D., Wehenkel, L.: Extremely randomized trees. Machine learning  \textbf{63},  3--42 (2006)

\bibitem{guidotti2022counterfactual}
Guidotti, R.: Counterfactual explanations and how to find them: literature review and benchmarking. Data Mining and Knowledge Discovery pp. 1--55 (2022)

\bibitem{lundberg2020local}
Lundberg, S.M., Erion, G., Chen, H., DeGrave, A., Prutkin, J.M., Nair, B., Katz, R., Himmelfarb, J., Bansal, N., Lee, S.I.: From local explanations to global understanding with explainable ai for trees. Nature machine intelligence  \textbf{2}(1),  56--67 (2020)

\bibitem{lundberg2017unified}
Lundberg, S.M., Lee, S.I.: A unified approach to interpreting model predictions. Advances in neural information processing systems  \textbf{30} (2017)

\bibitem{meinshausen2010node}
Meinshausen, N.: Node harvest. The Annals of Applied Statistics pp. 2049--2072 (2010)

\bibitem{meinshausen2006}
Meinshausen, N., Ridgeway, G.: Quantile regression forests. Journal of machine learning research  \textbf{7}(6) (2006)

\bibitem{molnar2022}
Molnar, C.: Interpretable Machine Learning. 2 edn. (2022)

\bibitem{scikit-learn}
Pedregosa, F., Varoquaux, G., Gramfort, A., Michel, V., Thirion, B., Grisel, O., Blondel, M., Prettenhofer, P., Weiss, R., Dubourg, V., Vanderplas, J., Passos, A., Cournapeau, D., Brucher, M., Perrot, M., Duchesnay, E.: Scikit-learn: Machine learning in {P}ython. Journal of Machine Learning Research  \textbf{12},  2825--2830 (2011)

\bibitem{ribeiro2016should}
Ribeiro, M.T., Singh, S., Guestrin, C.: " why should i trust you?" explaining the predictions of any classifier. In: Proceedings of the 22nd ACM SIGKDD international conference on knowledge discovery and data mining. pp. 1135--1144 (2016)

\bibitem{ribeiro2018anchors}
Ribeiro, M.T., Singh, S., Guestrin, C.: Anchors: High-precision model-agnostic explanations. In: Proceedings of the AAAI conference on artificial intelligence. vol.~32 (2018)

\bibitem{tan2020tree}
Tan, S., Soloviev, M., Hooker, G., Wells, M.T.: Tree space prototypes: Another look at making tree ensembles interpretable. In: Proceedings of the 2020 ACM-IMS on foundations of data science conference. pp. 23--34 (2020)

\bibitem{wu2018moleculenet}
Wu, Z., Ramsundar, B., Feinberg, E.N., Gomes, J., Geniesse, C., Pappu, A.S., Leswing, K., Pande, V.: Moleculenet: A benchmark for molecular machine learning (2018)

\end{thebibliography}
\bibliographystyle{splncs04}

\end{document}